# NeuroBridge: Bridging Multi-Task MRI Knowledge for Neurodegenerative Disease Diagnosis

**Mengyu Li[1,2,3], Guoyao Shen[1,2,3], Chad W. Farris[3,4,5*], Xin Zhang[1,2,3,6,7,8*]**

[1]Department of Mechanical Engineering, Boston University, Boston, MA 02215, United States

[2]The Photonics Center, Boston University, Boston, MA 02215, United States

[3]Rafik B. Hariri Institute for Computing and Computational Science & Engineering, Boston University, Boston, MA, 02215, United States

[4]Department of Radiology, Boston University Chobanian & Avedisian School of Medicine, Boston, MA 02118, United States

[5]Department of Radiology, Boston Medical Center, Boston, MA 02118, United States

[6]Department of Electrical and Computer Engineering, Boston University, Boston, MA 02215, United States

[7]Department of Biomedical Engineering, Boston University, Boston, MA 02215, United States

[8]Division of Materials Science and Engineering, Boston University, Boston, MA 02215, United States

*** Correspondence:**

Prof. Xin Zhang
xinz@bu.edu

Dr. Chad Farris
cfarris@bu.edu

**Keywords:** deep learning, transformer, multi-task learning, representation learning, neurodegenerative disease, brain MRI, opportunistic screening, dementia diagnosis.

**Abstract**

Accurate MRI-based identification of Alzheimer's disease (AD), mild cognitive impairment (MCI), and related dementias remains challenging because disease-related structural changes are often subtle and heterogeneous. We developed NeuroBridge, a clinically guided multi-task MRI framework for neurodegenerative disease diagnosis. NeuroBridge integrates large-scale self-supervised MRI pretraining with hippocampal segmentation, hippocampal atrophy classification, and reconstruction objectives, followed by gated fusion fine-tuning. Performance was evaluated across ADNI and OASIS cohorts, including cross-cohort transfer, probability-based analysis, and opportunistic screening. NeuroBridge achieved the highest performance across evaluated classification tasks, reaching 88.17% accuracy for AD versus cognitively normal controls in ADNI and 82.78% in OASIS. The largest gains occurred in MCI-related and mixed-diagnosis settings. The framework demonstrated strong cross-cohort generalization, systematic associations between predicted-class probability and accuracy, and the feasibility of probability-based opportunistic screening. Clinically guided multi-task representation learning improves neurodegenerative MRI diagnosis beyond conventional single-task approaches. NeuroBridge provides a robust and scalable framework for dementia assessment and MRI-based opportunistic screening.

## 1 Introduction

Neurodegenerative diseases, particularly Alzheimer's disease (AD), represent a growing global health and economic burden as populations age (Aranda et al., 2021). AD is characterized by a prolonged preclinical and prodromal phase during which subtle structural brain changes precede overt cognitive decline (Jack et al., 2018; Sperling et al., 2011). Mild cognitive impairment (MCI) is an intermediate stage between cognitively normal status (CN) and dementia and is associated with a substantially increased risk of progression to dementia (Gauthier et al., 2006; Petersen et al., 1999). Timely and accurate identification of individuals in early stages of neurodegenerative diseases, as well as reliable differentiation among MCI, AD, and other dementia subtypes, is critical for early intervention, treatment planning, and clinical trial enrollment (Frederiksen et al., 2024). The emergence of disease-modifying therapies, including anti-amyloid treatments, further elevates the importance of early identification, while evidence from oncology demonstrates the benefits of early detection for improving outcomes (Sun, 2026). However, delays in diagnosis and inability to predict neuropathology driving cognitive decline, particularly at the MCI stage, are well documented and underscore the need for new techniques to improve timely and more reliable identification (Mattke et al., 2023; Chen et al., 2024; Lang et al., 2017; Kusoro et al., 2025).

Recent advances in representation learning have introduced foundation-style models pretrained on large, heterogeneous datasets to capture generalizable structural priors (Bommasani et al., 2021; Moor et al., 2023). Masked autoencoders (MAEs), originally developed in natural image domains, have shown strong capability in learning transferable visual embeddings through self-supervised reconstruction objectives (He et al., 2022). When applied to medical imaging, such models can leverage millions of unlabeled slices to learn robust structural features before being adapted to downstream clinical tasks. Nevertheless, most medical imaging models are fine-tuned for individual tasks, limiting integration of complementary task-specific knowledge (Schäfer et al., 2024; Pai et al., 2024; Tak et al., 2026).

During clinical MRI interpretation, neurodegenerative diagnosis is rarely derived from a single source of knowledge. Instead, radiologists and neurologists integrate multiple cues, such as structural anatomy, region-specific degeneration patterns, and overall brain integrity, to form a unified diagnosis/impression (Frisoni et al., 2010; Wahlund et al., 2017). This workflow reflects a combination of complementary perspectives that can be mapped to different AI imaging tasks. Translating this multi-perspective reasoning into computational models remains challenging because individual learning objectives typically capture only isolated aspects of brain MRI. The resulting complementary representations must therefore be effectively integrated into a cohesive diagnostic signal (Lu et al., 2025).

Motivated by this observation, we introduce NeuroBridge, a transformer-based framework (Vaswani et al., 2017) that translates clinical workflow into computational modeling. Rather than treating imaging tasks as independent objectives, NeuroBridge is designed to integrate multiple MRI-derived findings into a unified diagnostic representation. The framework first learns heterogeneous anatomical representations through large-scale self-supervised MAE pretraining. These representations are subsequently enriched through multi-task pretraining, capturing complementary structural cues, including anatomical localization, degeneration patterns, and global structural information. In the final fine-tuning stage, NeuroBridge adaptively integrates these pretrained representations via a gated fusion mechanism, enabling coordinated utilization of task-specific features for downstream diagnosis. This design systematically bridges clinical workflow with computational modeling, linking anatomical representations to clinical insights and consolidating multi-task features into a unified diagnostic prediction.

Beyond classification performance alone, successful clinical deployment of neurodegenerative AI systems requires robustness across heterogeneous patient populations and meaningful assessment of prediction probabilities (Yang et al., 2024; Chua et al., 2023). Models that can identify both high-probability and ambiguous cases may better support real-world clinical workflows and facilitate downstream applications (Ianni et al., 2020; Young et al., 2021). Among these applications, opportunistic screening represents an especially attractive potential initial use case for NeuroBridge. Opportunistic screening refers to the detection of clinically important pathologies on imaging studies acquired for indications unrelated to the disease being identified (Pickhardt et al., 2023; Torisson et al., 2015; Bruno et al., 2025; Håkansson et al., 2019). This paradigm has already demonstrated clinical value in other imaging applications, including osteoporosis and cardiovascular disease, where opportunistic assessment has helped reduce delays in diagnosis and facilitate earlier intervention (Dagan et al., 2020; Sandhu et al., 2023; Pickhardt et al., 2020). Extending this concept to neurodegenerative disease may enable routine brain MRI examinations to serve as a scalable source of information for earlier identification of individuals at risk for cognitive impairment.

We therefore evaluate NeuroBridge across multiple neurodegenerative diagnostic tasks, cross-cohort transfer settings, and probability-based screening scenarios using the ADNI and OASIS cohorts. Results demonstrate improved diagnostic performance, enhanced generalization across datasets, and informative probability patterns that may support risk stratification.

# 2 Materials and Methods

## 2.1 Dataset details

To construct a robust and transferable structural 2D MRI encoder, we leveraged large-scale MRI datasets from multiple cohorts, including the National Alzheimer's Coordinating Center (NACC), the Alzheimer's Disease Neuroimaging Initiative (ADNI), Open Access Series of Imaging Studies (OASIS), and an additional radiology (RAD) dataset (Mei et al., 2022). These datasets collectively provide diverse acquisition protocols, scanner vendors, and demographic distributions, enabling broad structural representation learning. In total, the combined cohort for MAE encoder pretraining includes over 31 million 2D MRI slices.

Following MAE pretraining, the architecture is forked into multiple downstream decoders and further pretrained using supervised multi-task objectives that capture clinically relevant anatomical and pathological signals. The multi-task pretraining stage comprised hippocampal atrophy classification, hippocampal segmentation, and MAE reconstruction; the detailed methodology is described in Section 2.4.1. All three task-specific pretraining datasets were derived from the NACC dataset and only T1-weighted MR images were included. In particular, the target masks for hippocampal segmentation were generated from FreeSurfer software and hippocampal atrophy samples were manually extracted and labeled by the lead author (M.L.) under the guidance of a board-certified neuroradiologist (C.W.F.) with 6 years of dedicated neuroradiology practice. The dataset details for MAE encoder and multi-task pretraining can be found in Supplementary Material Table S1.

For downstream dementia classification, NeuroBridge was evaluated on the ADNI and OASIS cohorts. ADNI subjects were categorized as Alzheimer's disease (AD), cognitively normal (CN), early mild cognitive impairment (EMCI), and late mild cognitive impairment (LMCI). Following the ADNI data update released on March 26, 2026, all subjects with quality-control concerns (PTIDs formatted as 381_S_#####, corresponding to known image-processing issues) were excluded. OASIS subjects were categorized as AD, CN, AD with secondary diagnosis, other dementia, and Clinical Dementia Rating (CDR) scores of 0 and 0.5, representing cognitively normal and mild cognitive impairment populations, respectively. The AD with secondary diagnosis and other dementia groups enabled evaluation in the presence of mixed and non-AD dementias. Demographic characteristics of both cohorts are summarized in Table 1. These datasets offer complementary diagnostic distributions for both within-cohort evaluation and cross-cohort generalization experiments.

**Table 1.** Demographic and imaging characteristics of participants included in the downstream dementia classification tasks. Age is reported as mean ± standard deviation. Sex is reported as the number of male/female participants, and subject and scan counts are provided for each diagnostic group. The OASIS CDR 0 and CDR 0.5 groups were defined independently according to Clinical Dementia Rating scores. Abbreviations: CDR, Clinical Dementia Rating; CN, cognitively normal; EMCI, early mild cognitive impairment; LMCI, late mild cognitive impairment.

| ADNI | | | | |
|---|---|---|---|---|
| **Diagnostic group** | **AD** | **EMCI** | **LMCI** | **CN** |
| Age (years) | 75.60±7.93 | 73.57±7.68 | 75.06±7.53 | 75.93±6.83 |
| Sex (male/female) | 228/180 | 195/225 | 422/267 | 255/286 |
| Subjects | 408 | 420 | 689 | 541 |
| Scans | 558 | 636 | 883 | 3931 |

| OASIS | | | | |
|---|---|---|---|---|
| **Diagnostic group** | **AD** | **AD (Secondary Diagnosis)** | **Other Dementia** | **CN** |
| Age (years) | 74.56±7.20 | 72.68±7.52 | 71.17±7.57 | 66.65±8.93 |
| Sex (male/female) | 167/144 | 51/38 | 141/107 | 371/515 |
| Subjects | 311 | 89 | 248 | 886 |
| Scans | 444 | 112 | 347 | 2398 |
| **Diagnostic group** | **CDR 0** | | **CDR 0.5** | |
| Age (years) | 66.25±9.04 | | 73.99±7.03 | |
| Sex (male/female) | 452/302 | | 117/93 | |
| Subjects | 754 | | 210 | |
| Scans | 2241 | | 405 | |

## 2.2 Image preprocessing

The MRI datasets used in this study were acquired from multiple institutions and scanners, resulting in variability in voxel resolution, spatial orientation, field-of-view, and intensity distributions. To ensure consistent anatomical representation across datasets and facilitate large-scale multi-cohort learning, a standardized preprocessing pipeline was applied to all MRI scans.

All scans were registered to the 1-mm FSL MNI152 T1-weighted standard template using FSL (Andersson et al., 2007; Jenkinson et al., 2012) to establish consistent spatial alignment and anatomical correspondence across subjects. Following registration, the volumes were resampled to a uniform voxel resolution and standardized to dimensions of 256 × 256 × 256 through cropping or padding, ensuring compatibility across cohorts and stable batch processing during training. Finally, voxel intensities were normalized to the range of 0–1 using min–max normalization with 0.1%–99.9% clamping.

### 2.3 Model input construction and experimental design

During the multi-task pretraining stage, to leverage the 2D MAE pretraining encoder while preserving volumetric context, each 3D MRI volume was decomposed into ordered coronal 2D slices along the anterior–posterior axis. These slices were organized into a structured 2.5D representation of size $(B \times D \times H \times W)$, where $B$ is the batch size, $D$ denotes the number of slices per volume, and $H \times W$ represents the in-plane resolution. This representation preserves inter-slice ordering while enabling efficient slice-wise transformer encoding. During training, slices from the same subject remained grouped to maintain subject-level consistency.

All classification experiments employed subject-level data splits to prevent information leakage across scans of the same individual. For each dementia prediction task, instead of conventional $k$-fold cross-validation, we adopt Monte Carlo cross-validation, also known as repeated random sub-sampling validation, to reduce dependence on any single data partition while maintaining balanced validation and test sets (Varoquaux et al., 2017; Shan, 2022; Bradshaw et al., 2023). In this approach, the dataset is randomly partitioned into 60% training, 15% validation, and 25% test subsets 10 times, while preserving a 50/50 positive/negative class balance at each validation and test split. All remaining samples are assigned to the training set. An exception is made for the *AD with Secondary Diagnosis vs CN* task, where severe class imbalance (89 positive vs 886 negative) precludes balanced splitting; in this case, a single split of 85% training and 15% testing is adopted without a separate validation set.

To address inherent class imbalance in the training data, we further apply a weighted random sampling strategy during training, where sampling probabilities are inversely proportional to class frequencies. Consequently, samples from underrepresented classes were selected more often, producing approximately balanced class contributions across the training batches used for model updates. For each of the 10 Monte Carlo partitions, a new downstream model was initialized from the same pretrained checkpoint, trained independently, and evaluated on the corresponding test set. Final performance metrics are reported as the average and distribution across all Monte Carlo runs. For cross-cohort transfer experiments, models trained on one dataset (e.g., ADNI) were evaluated on another (e.g., OASIS) without retraining, enabling assessment of generalization across demographic and acquisition differences.

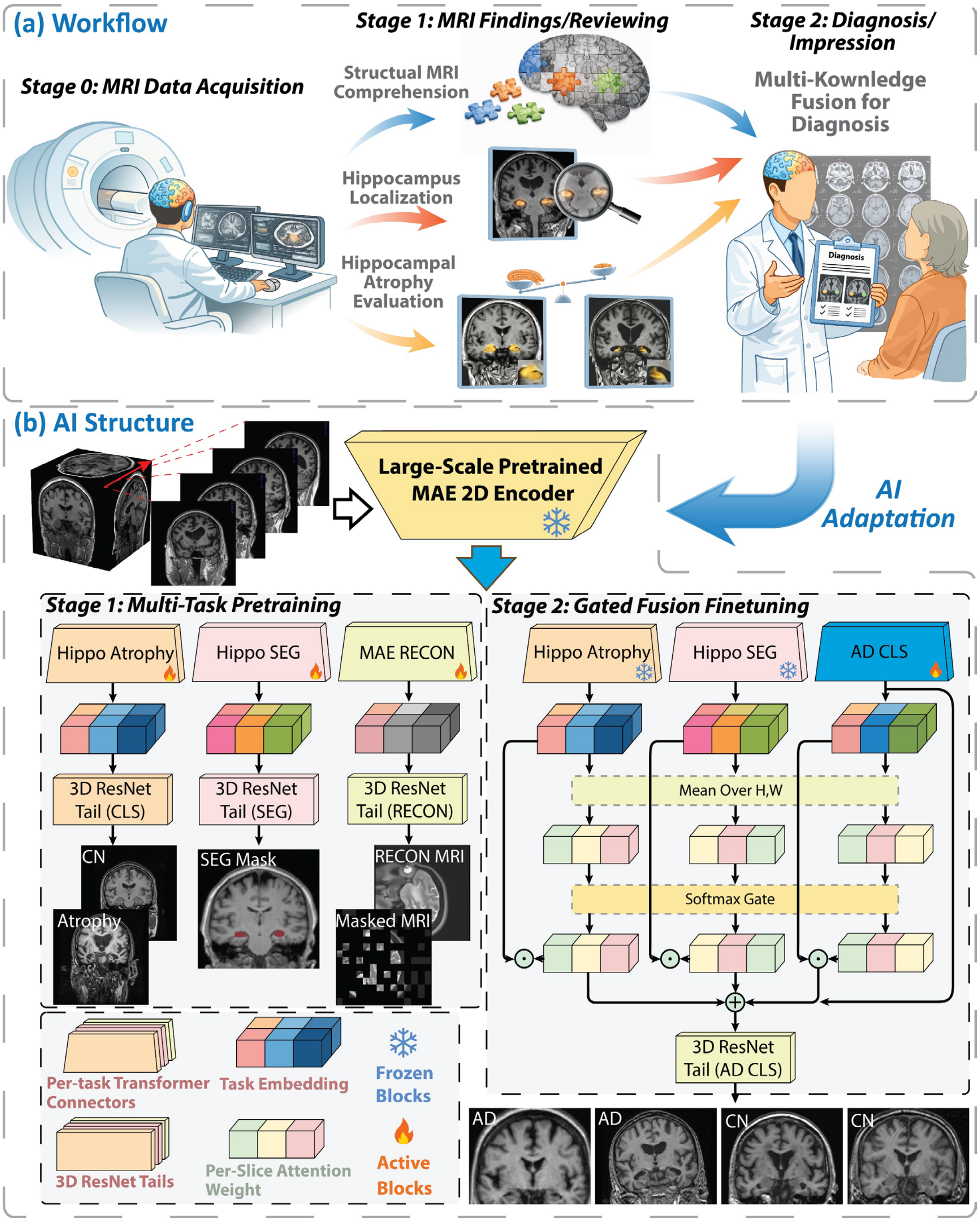

**Figure 1.** Overview of the NeuroBridge framework and its clinical-to-AI mapping. **(a)** Clinical workflow illustrating MRI-based neurodegenerative diagnosis. Radiological assessment integrates multi-source MRI findings (Stage 1), including structural MRI comprehension, hippocampal localization, and hippocampal atrophy evaluation, to form a unified diagnostic impression (Stage 2). **(b)** Corresponding AI architecture. A large-scale pretrained MAE 2D encoder provides shared anatomical representations. In Stage 1, multi-task pretraining learns complementary knowledge from hippocampal atrophy classification, hippocampal segmentation, and reconstruction tasks. In Stage 2, these pretrained features are adaptively fused via a gated mechanism, where slice-wise representations are aggregated and integrated for final dementia classification using a 3D ResNet head (demonstrated using AD classification (CLS) task). Human illustrations are schematic and are not intended to represent individual study participants or the demographic composition of the study cohorts.

## 2.4 NeuroBridge architecture

NeuroBridge is a two-stage framework that integrates large-scale self-supervised anatomical representation with clinically guided multi-task knowledge for neurodegenerative MRI diagnosis. As illustrated in Figure 1, the architecture is designed to reflect the clinical workflow of MRI-based assessment. In clinical practice (Figure 1a), radiological interpretation follows a two-stage process in which complementary MRI findings are first identified (Stage 1 in Figure 1a) and subsequently integrated into a unified diagnostic impression (Stage 2 in Figure 1a). NeuroBridge translates this paradigm into a computational framework (Figure 1b) by learning complementary representations through multi-task pretraining (Stage 1 in Figure 1b), followed by their integration via a gated fusion mechanism for downstream diagnosis (Stage 2 in Figure 1b). Both stages share a common MAE-based encoder, ensuring that task-specific representations are learned within a consistent feature space and can be effectively integrated for diagnosis. The training and modeling details for AI structure are illustrated in the following sections.

### 2.4.1 Stage 1: Multi-task pretraining

In the first stage, a 2D MAE transformer pretrained on large-scale multi-cohort MRI data is adopted as a shared encoder to provide generalizable anatomical representations. We followed the shared encoder pretraining protocol outlined in Li et al. (2026); additional details are provided in the Supplementary Material Section S1. The encoder is subsequently extended through a multi-task pretraining strategy incorporating hippocampal atrophy classification, hippocampal segmentation, and MAE reconstruction objectives. For each task, slice-wise encoder tokens are processed through task-specific transformer connectors and corresponding 3D ResNet tails. To ensure that global anatomical features are explicitly preserved and utilized for downstream classification, the MAE reconstruction branch additionally employs a dedicated MAE decoder while retaining its task-specific connector and 3D ResNet reconstruction tail. This design enables the model to learn complementary representations reflecting clinically relevant perspectives in MRI assessment, including degeneration patterns, anatomical localization, and global structural information. During this stage, the task-specific connectors, 3D ResNet tails, and MAE decoder are optimized while the shared encoder remains frozen to preserve its pretrained representations and ensure stable multi-task training.

When training with all tasks simultaneously, the model was optimized using AdamW optimizer (Loshchilov and Hutter, 2019) with a learning rate of $5 \times 10^{-4}$. The training objective combined hippocampal segmentation, hippocampal atrophy classification, and MAE reconstruction losses within a unified multi-task trainer. Hippocampal segmentation was formulated as a binary volumetric segmentation task, with the segmentation head producing single-channel logits and optimized using a segmentation loss combining focal binary cross-entropy ($\mathcal{L}_{FBCE}$) and log-cosh Dice ($\mathcal{L}_{LCDice}$) components:

$$\mathcal{L}_{seg} = \mathcal{L}_{FBCE} + \mathcal{L}_{LCDice} \tag{1}$$

The focal binary cross-entropy term is given by:

$$\mathcal{L}_{FBCE} = -w_{+}(1-p)^{\gamma} y \log(p) - w_{-} p^{\gamma}(1-y)\log(1-p) \tag{2}$$

where $p$ denotes the predicted probability, $y \in \{0,1\}$ is the ground-truth label, $\gamma = 2$ is the focusing parameter, and $w_{+} = 1.0$ and $w_{-} = 0.2$ are the weights for positive and negative voxels, respectively. The log-cosh Dice loss is defined as

$$D(p,y) = \frac{2\sum py + s}{\sum p^2 + \sum y^2 + s} \quad (3)$$

$$\mathcal{L}_{LCDice} = \frac{1}{N}\sum_{n=1}^{N} \log[\cosh(1 - D(p_n, y_n) + \epsilon)] \quad (4)$$

where $s = 1.0$ is the smoothing constant, $\epsilon = 10^{-12}$ is a small numerical offset inside the log-cosh operation, $n$ indexes samples in a batch and $N$ is the batch size. In this combined loss formulation, the focal term emphasizes difficult and underrepresented voxels, while the Dice term directly optimizes spatial overlap between the predicted and ground-truth segmentation masks.

The loss for hippocampal atrophy classification is formulated as the standard cross-entropy:

$$\mathcal{L}_{cls} = -\sum_{c} y_c \log(\hat{y}_c) \quad (5)$$

where $\hat{y}_c$ denotes the predicted class probability for class $c$, and $y_c$ is the corresponding ground-truth label.

The reconstruction loss follows the masked autoencoder objective and is computed only with masked patches:

$$\mathcal{L}_{recon} = \frac{1}{\sum_i m_i}\sum_{i} m_i \cdot \|\hat{x_i} - x_i\|^2 \quad (6)$$

where $\hat{x}$ and $x$ denote the predicted and target patch representations, respectively, and $m_i \in \{0,1\}$ indicates whether patch $i$ is masked.

The total training objective was formed by jointly optimizing the active task losses for each batch according to task identity. This setup enables coordinated optimization across tasks while preserving task-specific specialization within a shared representation space.

### 2.4.2 Stage 2: Gated Fusion Fine-tuning

In the second stage, NeuroBridge performs gated fusion fine-tuning for downstream dementia classification. Given a 2.5D input representation ($B \times D \times H \times W$), the shared MAE encoder produces slice-wise token embeddings, which are processed through pretrained connectors and a dementia classification branch initialized from the reconstruction branch. Rather than optimizing a single task-specific pathway, the framework integrates these representations through a slice-wise gated fusion mechanism shown in Figure 1b, Stage 2. Specifically, spatial features are aggregated via mean pooling over in-plane dimensions, and a softmax-based gating function is applied to generate normalized weights across branches for each slice. The resulting weighted features are fused into a unified volumetric representation and passed to a 3D ResNet classification head for final prediction.

During this stage, the pretrained encoder and task-specific connectors remain frozen, while only the gated fusion module and classification head are optimized. The model is trained using AdamW optimizer with a learning rate of $5 \times 10^{-5}$ and a weight decay of $1 \times 10^{-5}$. The training objective consists solely of dementia classification loss, formulated using the same cross-entropy loss as in Equation 5. This setup aims to enable coordinated reuse of multi-task pretrained representations, allowing complementary anatomical and pathological cues to be combined without retraining task-specific branches in Stage 1.

### 2.5 Probability-based threshold analysis

To evaluate whether model-output probability was associated with prediction correctness, the two output logits $z_{i,0}$, $z_{i,1}$ for each sample $i$ were converted to class probabilities using the softmax function:

$$p_{i,c} = \frac{e^{z_{i,c}}}{e^{z_{i,0}} + e^{z_{i,1}}}, \qquad c \in \{0, 1\} \tag{7}$$

The predicted class was determined as the class with the highest probability, and its corresponding probability was defined as:

$$p_i^{pred} = \max\left(p_{i,0}, p_{i,1}\right) \tag{8}$$

For each probability threshold $T$, only samples satisfying $p_i^{pred} \geq T$ were retained, and classification accuracy and retained sample fraction were calculated for this subset. Probability analysis was performed using the single best-performing model from each classification task. Positive-class probabilities were additionally used to visualize the distributions of model outputs across ground-truth classes.

For the exploratory opportunistic screening analysis, positive-class probabilities from the relevant disease-specific classifiers were compared with classifier-specific probability thresholds and combined to generate screening outcomes for cognitive abnormality. The threshold combinations and corresponding screening results are described in Section 3.5. These softmax-derived probabilities were used as model-output scores and were not interpreted as calibrated clinical risk estimates.

## 3 Results

### 3.1 Multi-task pretraining

We first evaluated the effectiveness of the multi-task pretraining stage, which integrates hippocampal segmentation, hippocampal atrophy classification, and masked autoencoding reconstruction objectives on the shared MAE encoder. Quantitative performance across tasks and representative qualitative results are summarized in Figure 2.

For the hippocampal segmentation task, the model achieved an intersection over union (IoU) of 58.48%, Dice score of 73.21%, precision of 72.65%, and recall of 74.47%. Qualitative examples in Figure 2a show that the predicted masks consistently localized the bilateral hippocampal regions and closely followed the corresponding anatomical boundaries, although minor differences remained at the mask margins. For the masked autoencoding reconstruction task, the model achieved a structural similarity index (SSIM) of 82.10% (Wang et al., 2004), mean squared error (MSE) of 0.43% (Wang and Bovik, 2009), and peak signal-to-noise ratio (PSNR) of 24.03 dB (Huynh-Thu and Ghanbari, 2008). Qualitative reconstruction examples in Figure 2b demonstrate that the MAE encoder successfully recovers global anatomical structure and fine-grained tissue contrast from masked inputs, preserving cortical and subcortical morphology.

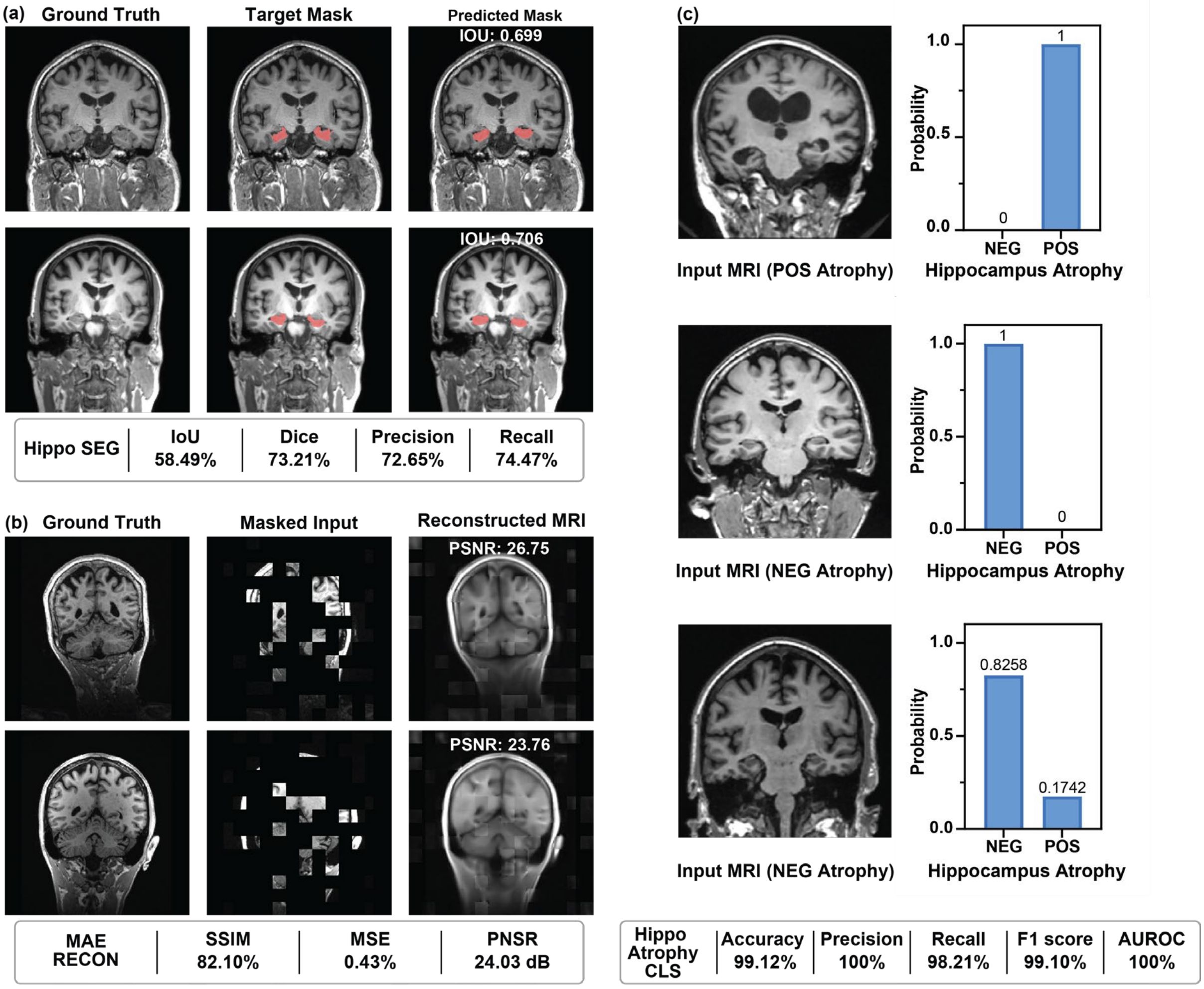

**Figure 2.** Representative and quantitative performance of NeuroBridge during multi-task pretraining. Representative results are shown for **(a)** hippocampal segmentation, **(b)** masked autoencoder reconstruction, and **(c)** hippocampal atrophy classification. Panel (a) compares the predicted hippocampal segmentation masks with the reference annotations (top), with overall performance summarized using IoU, Dice score, precision, and recall (bottom). Panel (b) compares the reconstructed MRI images with the original and masked inputs (top), with overall performance summarized using SSIM, MSE, and PSNR (bottom). Panel (c) presents representative hippocampal atrophy classification outputs (top), with overall performance summarized using accuracy, precision, recall, F1 score, and AUROC (bottom). All quantitative metrics are reported as percentages except PSNR, which is reported in decibels. *Abbreviations*: AUROC, area under the receiver operating characteristic curve; CLS, classification; IoU, intersection over union; MAE, masked autoencoder; MSE, mean squared error; PSNR, peak signal-to-noise ratio; RECON, reconstruction; SEG, segmentation; SSIM, structural similarity index measure.

In hippocampal atrophy classification, the model achieved an accuracy of 99.12%, with precision of 100%, recall of 98.21%, F1 score of 99.10%, and AUROC of 100%. Representative classification outputs are shown in Figure 2c, where predicted probabilities were obtained by applying a softmax function to the output logits (Equation 7). Together, these results summarize the performance of the three auxiliary objectives used for subsequent downstream fine-tuning.

## 3.2 Downstream dementia prediction

We next evaluated NeuroBridge on downstream dementia classification tasks across the ADNI and OASIS cohorts. Two NeuroBridge variants were examined: NeuroBridge (Seg), which incorporates hippocampal segmentation pretraining, and NeuroBridge (Seg+Hippo), which integrates both hippocampal segmentation and hippocampal atrophy classification representations. Because the reconstruction-pretrained pathway is used to initialize the downstream dementia classification pathway in all configurations, it is common to both NeuroBridge variants and is therefore omitted in their abbreviated names. DenseNet121 (Huang et al., 2017), MedViT (Manzari et al., 2023), ResNet2D (He et al., 2016), and ResNet3D (Hara et al., 2018) were included as conventional 2D and 3D baselines. All baseline models were trained independently using the same subject-level data partitions, preprocessing pipeline, sampling strategy, optimizer, learning rate, and weight decay used for Stage 2, as described in Section 2.4.2. Performance comparisons are summarized in Table 2, accuracy distributions across repeated experiments are shown in Figure 3, and raw NeuroBridge results for the individual Monte Carlo partitions are provided in Supplementary Material Tables S2 and S3.

### 3.2.1 ADNI cohort

Across all ADNI classification tasks, NeuroBridge (Seg+Hippo) achieved the highest overall performance among all evaluated models (Table 2 ADNI). For *AD vs CN* classification, NeuroBridge achieved 88.17% accuracy and 94.23% AUROC, exceeding the strongest baseline, ResNet3D, which achieved 84.91% accuracy and 93.60% AUROC. For *AD vs EMCI* and *CN vs LMCI*, NeuroBridge achieved the highest accuracies of 83.39% and 74.82%, respectively. The largest accuracy gain was observed for *AD vs LMCI* classification, where NeuroBridge achieved 73.32% accuracy compared with 64.58% for ResNet3D, corresponding to an 8.74 percentage-point improvement. *For CN vs EMCI* classification, NeuroBridge achieved the highest accuracy (75.13%), F1 score (81.15%), and AUROC (80.56%).

### 3.2.2 OASIS cohort

Across OASIS classification tasks, NeuroBridge (Seg+Hippo) achieved the strongest overall performance (Table 2 OASIS). It reached 82.78% accuracy, 79.33% F1 score, and 89.69% AUROC for *AD vs CN*, and also obtained the highest accuracy and AUROC for *AD with Secondary Diagnosis vs CN*, *AD + AD with Secondary Diagnosis vs CN*, *CDR 0 vs CDR 0.5*, and *Other Dementia vs CN*. Notably, it achieved 83.07% accuracy and 90.72% AUROC for *CDR 0 vs CDR 0.5*, and 80.40% accuracy and 85.19% AUROC for *Other Dementia vs CN*. The largest OASIS accuracy gain was observed for *CDR 0 vs CDR 0.5*, where NeuroBridge achieved 83.07% accuracy compared with 74.87% for the strongest baseline, ResNet3D, corresponding to an 8.20 percentage point improvement.

**Table 2.** Classification performance across ADNI and OASIS neurodegenerative diagnostic tasks. Performance is reported across five ADNI tasks (*AD vs CN*, *AD vs LMCI*, *AD vs EMCI*, *CN vs EMCI*, and *CN vs LMCI*) and five OASIS tasks (*AD vs CN, AD with secondary diagnosis vs CN*, *AD + AD with Secondary Diagnosis vs CN*, *CDR 0 vs CDR 0.5*, and *other dementia vs CN*). Values are mean accuracy, precision, recall, F1 score, and AUROC across 10 Monte Carlo partitions. Abbreviations: CDR, Clinical Dementia Rating; CN, cognitively normal; EMCI, early mild cognitive impairment; LMCI, late mild cognitive impairment; Seg, hippocampal segmentation pretraining; Seg+Hippo, hippocampal segmentation and hippocampal atrophy classification pretraining.

| Method | ADNI | | | | |
|---|---|---|---|---|---|
| | **Accuracy (%)** | **Precision (%)** | **Recall (%)** | **F1 Score (%)** | **AUROC (%)** |
| | | | **AD vs CN** | | |
| DenseNet121 | 82.55 | 91.46 | 71.11 | 79.48 | 91.64 |
| MedViT | 83.56 | 89.52 | 75.37 | 81.32 | 91.19 |
| ResNet2D | 84.60 | 90.60 | 77.69 | 83.02 | 92.58 |
| ResNet3D | 84.91 | 91.55 | 77.76 | 82.96 | 93.60 |
| *NeuroBridge(Seg)* | 86.55 | 87.76 | 82.31 | 84.86 | 93.30 |
| *NeuroBridge(Seg+Hippo)* | **88.17** | **91.69** | **83.13** | **87.20** | **94.23** |
| | | | **AD vs LMCI** | | |
| DenseNet121 | 61.58 | 68.53 | 60.79 | 63.37 | 65.71 |
| MedViT | 63.68 | 70.31 | 62.60 | 65.24 | 67.79 |
| ResNet2D | 64.02 | 69.84 | 67.57 | 67.98 | 71.14 |
| ResNet3D | 64.58 | 70.95 | 68.62 | 69.08 | 72.16 |
| *NeuroBridge(Seg)* | 68.97 | 78.95 | 63.45 | 70.35 | 77.05 |
| *NeuroBridge(Seg+Hippo)* | **73.32** | **80.76** | **70.85** | **75.29** | **79.02** |
| | | | **AD vs EMCI** | | |
| DenseNet121 | 76.53 | 86.28 | 71.23 | 77.05 | 87.35 |
| MedViT | 75.16 | 83.30 | 72.94 | 76.41 | 85.05 |
| ResNet2D | 76.98 | 88.35 | 69.79 | 77.52 | 87.86 |
| ResNet3D | 77.70 | **89.12** | 70.84 | 77.64 | 89.49 |
| *NeuroBridge(Seg)* | 80.08 | 86.54 | 72.31 | 78.57 | 88.81 |
| *NeuroBridge(Seg+Hippo)* | **83.39** | 88.88 | **81.54** | **84.94** | **90.55** |
| | | | **CN vs EMCI** | | |
| DenseNet121 | 70.13 | 68.69 | 69.84 | 68.75 | 71.18 |
| MedViT | 71.76 | 72.21 | 70.27 | 70.82 | 72.65 |
| ResNet2D | 72.12 | 70.60 | 74.89 | 72.51 | 73.94 |
| ResNet3D | 72.34 | 71.06 | 70.16 | 70.35 | 74.46 |
| *NeuroBridge(Seg)* | 72.60 | **78.78** | 75.22 | 75.94 | 80.38 |
| *NeuroBridge(Seg+Hippo)* | **75.13** | 73.76 | **90.99** | **81.15** | **80.56** |
| | | | **CN vs LMCI** | | |
| DenseNet121 | 66.19 | 80.14 | 57.29 | 60.97 | 78.64 |
| MedViT | 69.04 | **89.60** | 49.20 | 61.63 | 79.64 |
| ResNet2D | 69.13 | 86.91 | 53.39 | 63.18 | 78.85 |
| ResNet3D | 70.25 | 81.17 | 50.73 | 64.10 | 81.74 |
| *NeuroBridge(Seg)* | 73.04 | 80.14 | **66.80** | 72.86 | 78.64 |
| *NeuroBridge(Seg+Hippo)* | **74.82** | 85.72 | 64.32 | **73.39** | **83.50** |

*Continue to OASIS*

| Method | OASIS | | | | |
|---|---|---|---|---|---|
| | Accuracy (%) | Precision (%) | Recall (%) | $F_1$ Score (%) | AUROC (%) |
| | | | **AD vs CN** | | |
| DenseNet121 | 71.01 | 65.85 | 68.59 | 66.19 | 79.12 |
| MedViT | 72.38 | 66.80 | 71.58 | 68.25 | 80.39 |
| ResNet2D | 74.54 | 72.20 | 67.18 | 68.47 | 81.21 |
| ResNet3D | 74.68 | 70.63 | 70.34 | 69.70 | 82.28 |
| *NeuroBridge(Seg)* | 79.25 | 79.42 | 68.31 | 73.15 | 86.02 |
| *NeuroBridge(Seg+Hippo)* | **82.78** | **79.51** | **79.32** | **79.33** | **89.69** |
| | | | **AD with Secondary Diagnosis vs CN** | | |
| DenseNet121 | 77.22 | 55.71 | 58.21 | 56.93 | 82.29 |
| MedViT | 75.56 | 53.44 | 58.21 | 55.04 | 77.49 |
| ResNet2D | 76.33 | 54.30 | **63.88** | 57.85 | 81.24 |
| ResNet3D | 78.22 | 62.30 | 45.07 | 50.05 | 81.58 |
| *NeuroBridge(Seg)* | 76.02 | 54.80 | 60.00 | 56.46 | 80.51 |
| *NeuroBridge(Seg+Hippo)* | **80.77** | **66.70** | 51.63 | **57.85** | **81.75** |
| | | | **AD + AD with Secondary Diagnosis vs CN** | | |
| DenseNet121 | 77.64 | 58.43 | 57.31 | 56.47 | 81.57 |
| MedViT | 75.99 | 75.25 | 68.74 | 70.72 | 85.01 |
| ResNet2D | 76.63 | 70.99 | 79.42 | 74.69 | 85.82 |
| ResNet3D | 75.79 | 74.75 | 69.11 | 70.83 | 84.63 |
| *NeuroBridge(Seg)* | 79.42 | 81.82 | 68.00 | 74.28 | 88.19 |
| *NeuroBridge(Seg+Hippo)* | **84.31** | **82.47** | **81.36** | **81.86** | **90.26** |
| | | | **CDR 0 vs CDR 0.5** | | |
| DenseNet121 | 74.62 | 75.71 | 65.76 | 67.84 | 84.77 |
| MedViT | 70.08 | 59.70 | 74.19 | 65.01 | 79.76 |
| ResNet2D | 72.19 | 64.32 | 75.86 | 67.42 | 83.33 |
| ResNet3D | 74.87 | 66.01 | 77.80 | 70.27 | 84.29 |
| *NeuroBridge(Seg)* | 79.55 | 71.08 | **77.96** | 74.36 | 90.70 |
| *NeuroBridge(Seg+Hippo)* | **83.07** | **79.69** | 74.78 | **77.03** | **90.72** |
| | | | **Other Dementia vs CN** | | |
| DenseNet121 | 73.62 | 62.95 | 76.83 | 69.20 | 82.88 |
| MedViT | 74.72 | 67.54 | 75.16 | 70.35 | 82.06 |
| ResNet2D | 75.85 | 72.74 | 71.94 | 71.23 | 83.49 |
| ResNet3D | 76.11 | 71.46 | 74.27 | 71.82 | 83.19 |
| *NeuroBridge(Seg)* | 74.12 | 71.58 | 62.50 | 66.45 | 77.61 |
| *NeuroBridge(Seg+Hippo)* | **80.40** | **76.62** | **75.24** | **75.48** | **85.19** |

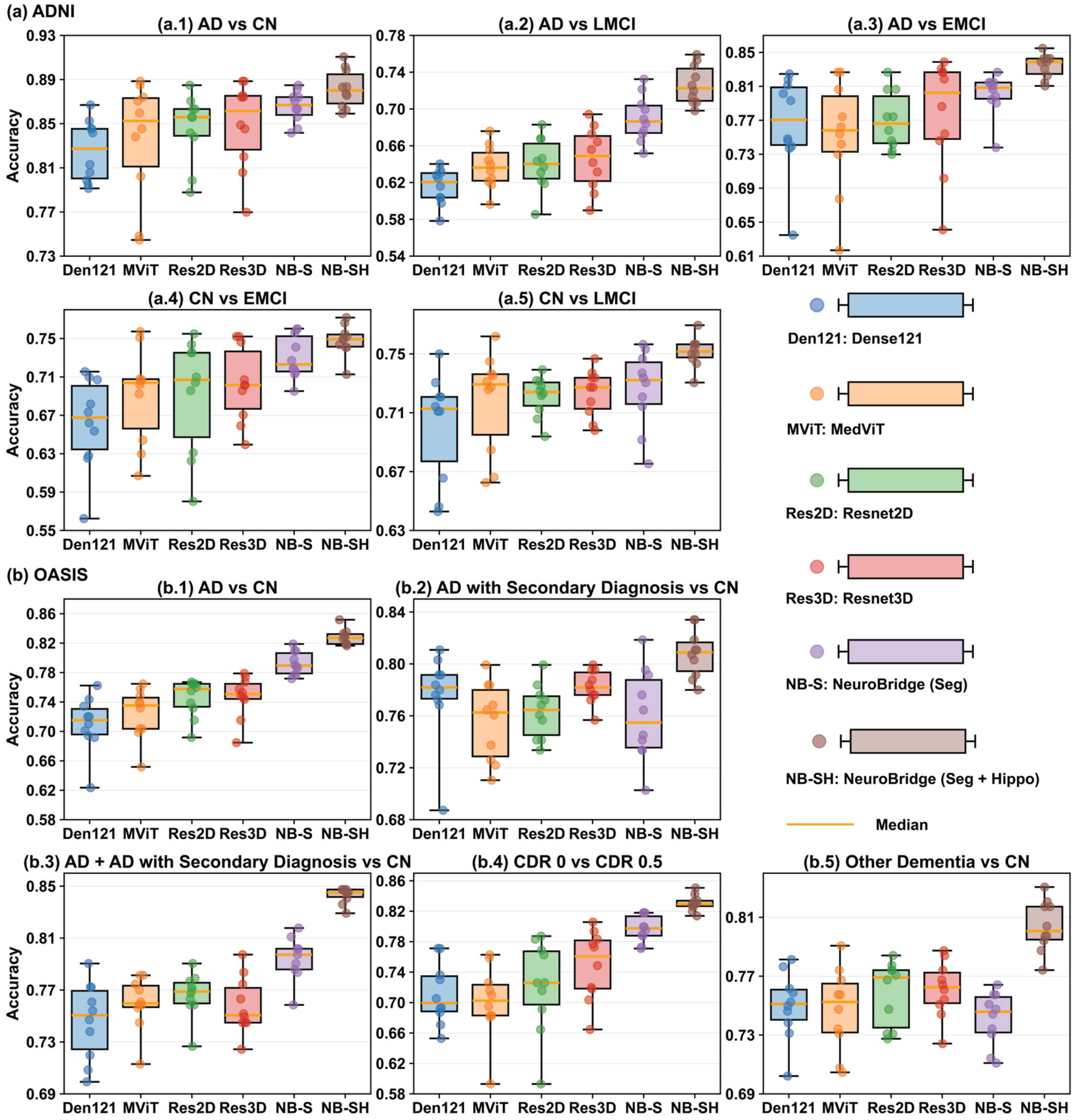

**Figure 3.** Accuracy distributions across ADNI and OASIS classification tasks. Boxplots compare model accuracy across repeated experiments for **(a)** ADNI and **(b)** OASIS classification tasks. ADNI tasks include (a.1) *AD vs CN*, (a.2) *AD vs LMCI*, (a.3) *AD vs EMCI*, (a.4) *CN vs EMCI*, and (a.5) *CN vs LMCI*. OASIS tasks include (b.1) *AD vs CN*, (b.2) *AD with secondary diagnosis vs CN*, (b.3) *AD + AD with secondary diagnosis vs CN*, (b.4) *CDR 0 vs CDR 0.5*, and (b.5) *other dementia vs CN*. Each overlaid point represents the accuracy obtained from one independently generated data partition for the corresponding classification task.

### 3.2.3 Performance consistency across repeated experiments

Figure 3 further demonstrates that NeuroBridge (Seg+Hippo) achieved consistent performance improvements across repeated experiments, with higher median accuracy, upward-shifted accuracy distributions, fewer low-performing runs, and generally tighter variability than the CNN and transformer baselines. The largest separations were observed in clinically challenging tasks involving neighboring disease stages, including *AD vs LMCI*, *CN vs EMCI*, and *CDR 0 vs CDR 0.5*, as well as heterogeneous dementia groups such as AD with secondary diagnoses and other dementia etiologies. Overall, NeuroBridge showed its most pronounced gains in MCI-related and mixed-diagnosis classifications across both ADNI and OASIS.

### 3.3 Cross-cohort generalization

To further evaluate robustness under real-world distribution shifts, we conducted cross-cohort transfer experiments in which, for each architecture and transfer direction, a single model trained on one cohort was evaluated directly on the other without retraining, fine-tuning, or domain adaptation. This setting assessed model generalization across differences in demographics, diagnostic composition, scanner protocols, and image acquisition conditions. Results are summarized in Table 3.

**Table 3.** Cross-cohort transfer performance for *AD vs CN* classification. Models trained on one cohort were evaluated on the other. NeuroBridge denotes NeuroBridge (Seg+Hippo), the fully pretrained configuration incorporating hippocampal segmentation and atrophy-classification representations. Performance is reported using accuracy, precision, recall, F1 score, and AUROC, expressed as percentages. Abbreviations: CN, cognitively normal; Seg+Hippo, hippocampal segmentation and hippocampal atrophy classification pretraining.

| **ADNI (Trained on OASIS)** | | | | | |
|---|---|---|---|---|---|
| **Method** | **Accuracy (%)** | **Precision (%)** | **Recall (%)** | **$F_1$ Score (%)** | **AUROC (%)** |
| | | | **AD vs CN** | | |
| DenseNet121 | 82.01 | 75.30 | 82.09 | 78.55 | 90.07 |
| MedViT | 80.17 | 78.22 | 68.11 | 72.82 | 85.45 |
| ResNet2D | 82.73 | 86.55 | 76.87 | 81.42 | 90.77 |
| ResNet3D | 81.65 | 86.72 | 73.13 | 79.35 | 88.25 |
| *NeuroBridge* | **88.13** | **89.14** | **85.82** | **87.45** | **91.48** |
| **OASIS (Trained on ADNI)** | | | | | |
| **Method** | **Accuracy (%)** | **Precision (%)** | **Recall (%)** | **$F_1$ Score (%)** | **AUROC (%)** |
| | | | **AD vs CN** | | |
| DenseNet121 | 80.48 | 83.11 | 69.49 | 75.69 | 87.26 |
| MedViT | 74.59 | 87.10 | 45.76 | 60.00 | 86.92 |
| ResNet2D | 81.41 | 75.78 | 71.35 | 73.50 | 88.97 |
| ResNet3D | 81.64 | 85.61 | 67.23 | 75.31 | 89.63 |
| *NeuroBridge* | **84.71** | **87.84** | **73.45** | **80.00** | **92.24** |

When trained on the *AD vs CN* task using OASIS and evaluated on ADNI, NeuroBridge achieved the highest performance across all metrics, reaching 88.13% accuracy, 89.14% precision, 85.82% recall, 87.45% F1 score, and 91.48% AUROC. Compared with the strongest conventional baseline (ResNet2D), NeuroBridge improved accuracy by 5.40 percentage points, recall by 8.95 percentage points, F1 score by 6.03 percentage points, and AUROC by 0.71 percentage points.

A similar trend was observed when models were trained on ADNI and evaluated on OASIS. NeuroBridge again achieved the highest overall performance, with 84.71% accuracy, 87.84% precision, 73.45% recall, 80.00% F1 score, and 92.24% AUROC. Relative to the strongest baseline (ResNet3D), NeuroBridge improved accuracy by 3.07 percentage points, recall by 6.22 percentage points, F1 score by 4.69 percentage points, and AUROC by 2.61 percentage points. Notably, NeuroBridge trained on ADNI achieved 84.71% accuracy on OASIS, exceeding the corresponding within-cohort OASIS accuracy of 82.78% reported in Table 2.

Across both transfer directions, NeuroBridge consistently achieved the highest evaluation metrics among all evaluated models.

### 3.4 Prediction probability evaluation

We further evaluated model behavior under varying predicted-class probability thresholds to assess the association between model-output probability and classification accuracy. To isolate probability-threshold behavior from variability introduced by repeated Monte Carlo runs, this analysis was performed using the single best-performing NeuroBridge model from each classification task. As shown in Figures 4a and 4c, classification accuracy and retained sample fraction, defined as the proportion of samples with predicted-class probabilities exceeding a specified threshold, were analyzed across the ADNI and OASIS cohorts.

Across ADNI tasks (Figure 4a), classification accuracy increased as probability thresholds became more stringent, while retained sample fractions decreased. *AD vs CN* and *AD vs EMCI* exhibited higher baseline accuracies and smaller threshold-dependent gains. In contrast, larger improvements were observed for *AD vs LMCI*, *CN vs EMCI*, and *CN vs LMCI*, where lower baseline accuracies increased substantially after low-probability samples were excluded.

The probability distributions in Figure 4b showed that many predictions were concentrated near positive-class probabilities of 0 or 1, whereas predictions with probabilities closer to 0.5 were located nearer the classification decision boundary. Compared with *AD vs CN* and *AD vs EMCI*, LMCI-related tasks exhibited broader distributions and greater overlap between positive and negative classes.

Similar trends were observed in the OASIS cohort (Figures 4c and 4d). Classification accuracy generally increased as the probability threshold became more stringent, while the retained sample fraction decreased. Compared with *AD vs CN*, *AD with Secondary Diagnosis vs CN* and *Other Dementia vs CN* tasks showed greater overlap between class-specific probability distributions, with more positive cases assigned low probabilities and more CN cases assigned elevated positive-class probabilities, indicating greater prediction ambiguity.

The retained-fraction curves demonstrate that improved accuracy was achieved across a substantial subset of the cohort rather than a small number of samples. For example, ADNI *AD vs CN* accuracy increased from 0.910 to approximately 0.96 while retaining 237 of 278 samples (85.3%), whereas OASIS *CDR 0 vs CDR 0.5* accuracy approached 0.96 while retaining 361 of 489 samples (73.8%). In contrast, more challenging tasks such as *AD vs LMCI* required a larger reduction in retained samples to achieve comparable accuracy gains. Overall, higher probability thresholds consistently selected subsets with improved classification accuracy across both ADNI and OASIS cohorts.

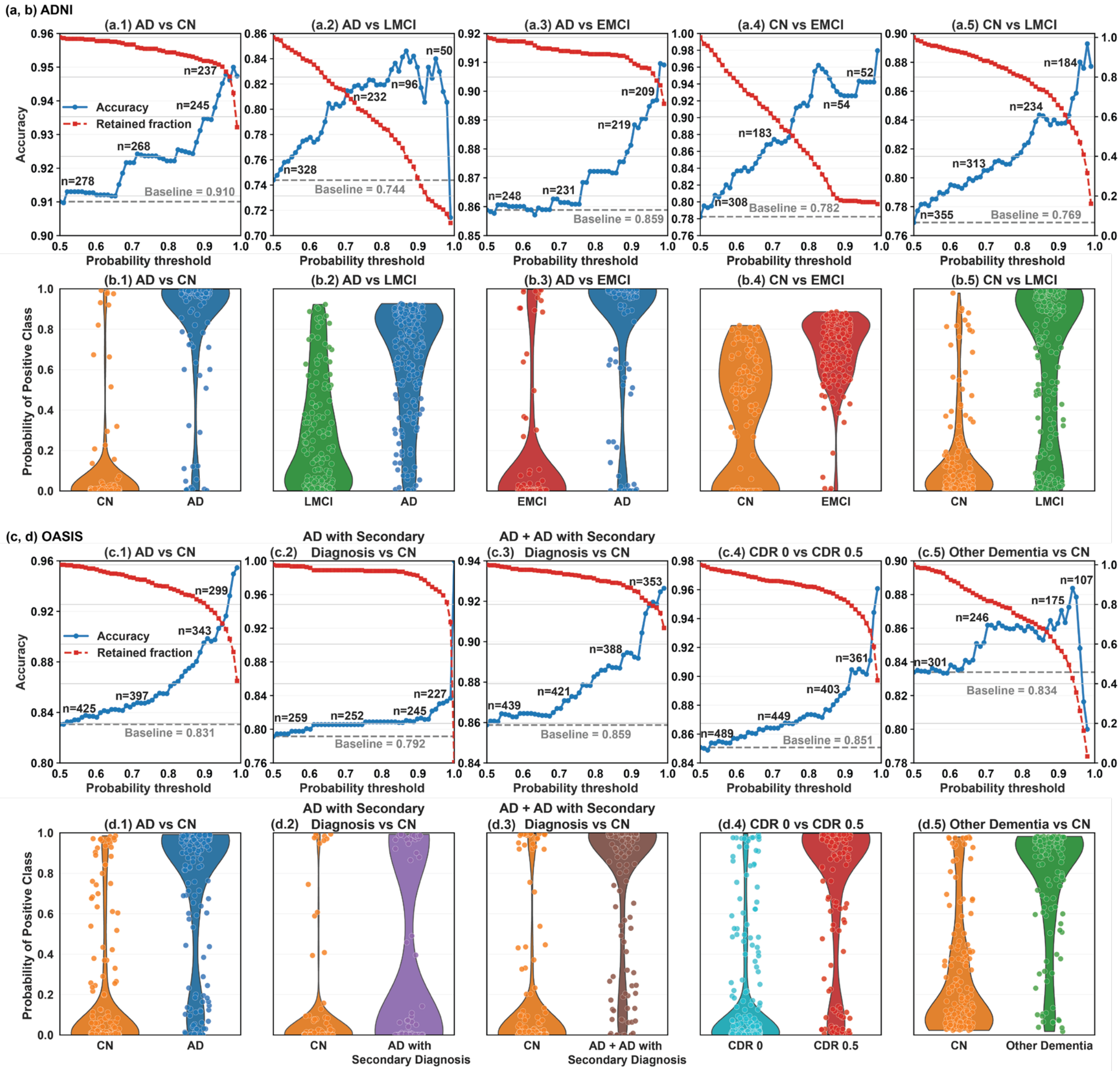

**Figure 4.** Classification performance under probability thresholding and model-output probability distributions across ADNI and OASIS. **(a–b)** ADNI results. (a) Accuracy and retained fraction as functions of probability threshold for five binary classification tasks: *AD vs CN*, *AD vs LMCI*, *AD vs EMCI*, *CN vs EMCI*, and *CN vs LMCI*. The blue curve denotes classification accuracy, the red dashed curve indicates the retained fraction of samples above each threshold, and the gray dashed line marks baseline accuracy using all samples. Sample counts at selected thresholds are annotated. (b) Corresponding distributions of positive-class probabilities for each task, shown as violin plots with overlaid data points, stratified by ground-truth class. **(c–d)** OASIS results. (c) Accuracy and retained fraction versus probability threshold for five tasks: *AD vs CN*, *AD with secondary diagnosis vs CN*, *AD + AD with secondary diagnosis vs CN*, *CDR 0 vs CDR 0.5*, and *other dementia vs CN*. (d) Corresponding positive-class probability distributions for each task.

### 3.5 Potential application of opportunistic screening

To explore a potential screening application of NeuroBridge, we constructed a retrospective simulated opportunistic screening cohort using a predefined patient-level holdout set from ADNI (Supplementary Material Table S4). This combined cohort was used to approximate a population of brain MRI examinations that could be encountered in a routine clinical setting. As illustrated in Figure 5a, each MRI examination was processed using multiple NeuroBridge classifiers, including *AD vs CN*, *LMCI vs CN*, *EMCI vs CN*, and *AD vs EMCI* models. Each classifier generated a positive-class probability, which was compared with an adjustable threshold (*T*). Predictions exceeding the specified threshold were treated as positive screening signals, whereas those below the threshold were treated as negative. The thresholded outputs from the relevant disease-specific classifiers were aggregated to produce two screening outcomes: (1) identification of all cognitively abnormal subjects (*EMCI + LMCI + AD vs CN*) and (2) identification of clinically significant impairment (*LMCI + AD vs EMCI + CN*). For case (1), prediction is positive when:

$$p_{AD\ vs\ CN} \geq T_{AD\ vs\ CN} \vee p_{LMCI\ vs\ CN} \geq T_{LMCI\ vs\ CN} \vee p_{EMCI\ vs\ CN} \geq T_{EMCI\ vs\ CN} \quad (9)$$

and for case (2), prediction is positive when:

$$(p_{AD\ vs\ CN} \geq T_{AD\ vs\ CN} \vee p_{LMCI\ vs\ CN} \geq T_{LMCI\ vs\ CN}) \wedge p_{AD\ vs\ EMCI} \geq T_{AD\ vs\ EMCI} \quad (10)$$

where $p_*$ and $T_*$ represent positive probability and threshold of target classifiers.

This retrospective simulation enabled evaluation of how positive-class probabilities from multiple classifiers could be integrated into a unified screening framework while providing explicit control over the sensitivity–specificity trade-off through threshold selection. Figures 5b and 5c summarize the effects of varying classifier probability thresholds on screening performance. Figure 5b visualizes the continuous performance landscape across threshold combinations, while Figure 5c highlights representative corner and intermediate operating points selected from the threshold-response landscapes shown in Figure 5b. Across both screening scenarios, increasing probability thresholds reduced sensitivity while increasing specificity and positive predictive value (PPV), producing a range of operating points with different performance characteristics.

For detection of all cognitively abnormal subjects (EMCI, LMCI, and AD) versus cognitively normal controls, the lowest threshold combination (Threshold = 0.50, 0.50, 0.50) achieved the highest sensitivity (85.21%), maximizing identification of abnormal individuals but with relatively low specificity (54.02%). In contrast, the strictest threshold combination (Threshold = 0.95, 0.95, 0.95) yielded near-ceiling specificity (98.20%) and PPV (98.23%), indicating that nearly all positive screening results corresponded to true cognitive abnormalities, although sensitivity decreased to 45.18%. An intermediate operating point (Threshold = 0.75, 0.75, 0.75) provided a more balanced performance profile, achieving 64.58% sensitivity, 86.86% specificity, and 91.53% PPV.

A similar trend was observed for detection of LMCI + AD versus EMCI + CN. The lowest threshold combination achieved the highest sensitivity (65.93%), favoring broader case detection, whereas the strictest threshold combination increased specificity to 95.46% and PPV to 91.39% while reducing sensitivity to 46.96%. The intermediate threshold setting provided a balance between detection sensitivity and false-positive control, achieving 61.14% sensitivity, 92.43% specificity, and 89.24% PPV.

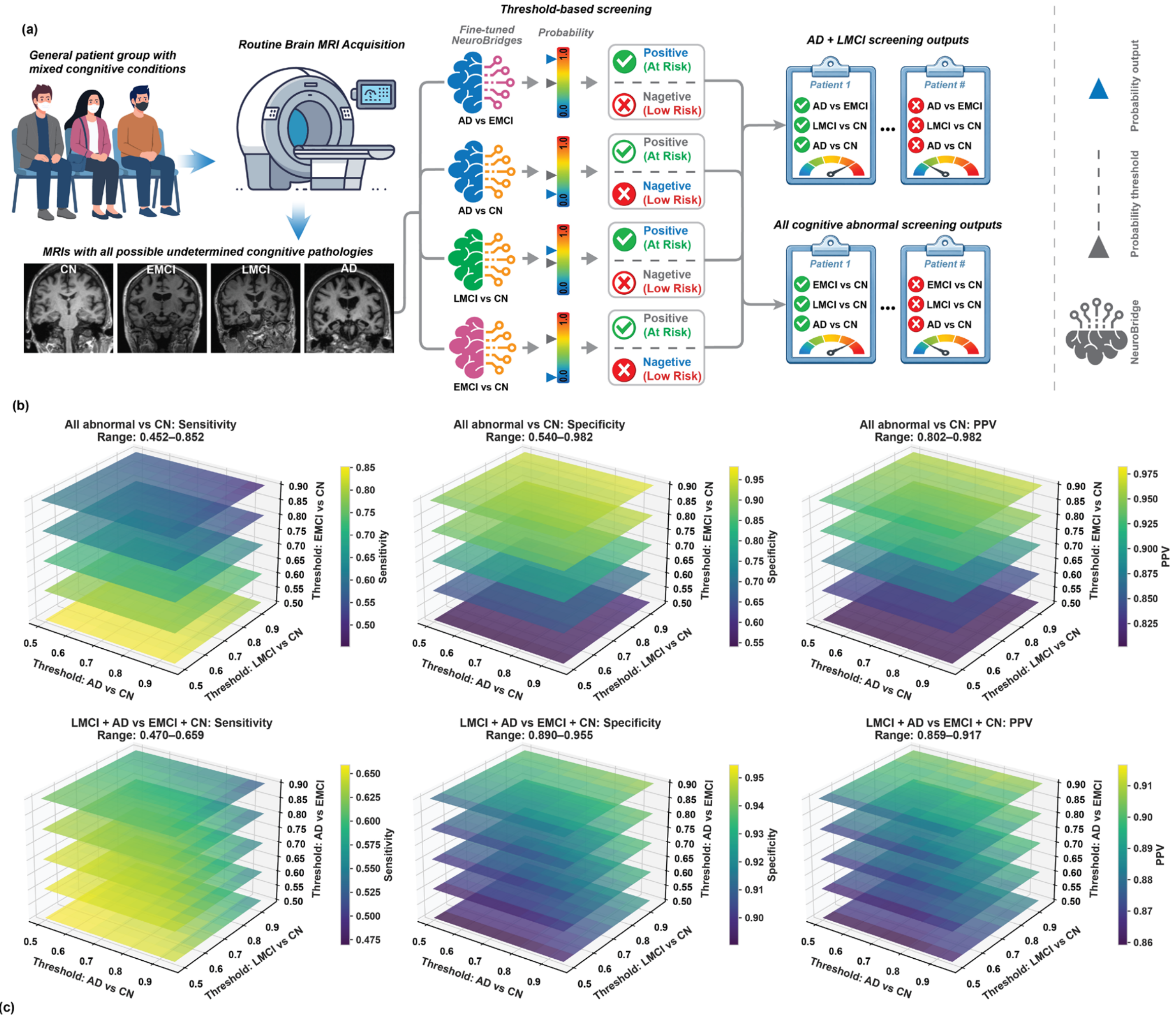

(c)

| All abnormal vs CN | | | |
|---|---|---|---|
| Threshold (AD vs CN, LMCI vs CN, EMCI vs CN) | Sensitivity | Specificity | PPV |
| (0.5, 0.5, 0.5) | **0.8521** | 0.5402 | 0.8030 |
| (0.95, 0.5, 0.5) | 0.8385 | 0.5462 | 0.8025 |
| (0.95, 0.95, 0.5) | 0.7652 | 0.6029 | 0.8091 |
| (0.5, 0.95, 0.5) | 0.8222 | 0.5880 | 0.8145 |
| (0.5, 0.5, 0.95) | 0.6295 | 0.8865 | 0.9243 |
| (0.95, 0.5, 0.95) | 0.5861 | 0.9014 | 0.9290 |
| (0.95, 0.95, 0.95) | 0.4518 | **0.9820** | **0.9823** |
| (0.5, 0.95, 0.95) | 0.5753 | 0.9552 | 0.9658 |
| (0.75, 0.75, 0.75) | 0.6458 | 0.8686 | 0.9153 |

| LMCI + AD vs EMCI + CN | | | |
|---|---|---|---|
| Threshold (AD vs CN, LMCI vs CN, AD vs EMCI) | Sensitivity | Specificity | PPV |
| (0.5, 0.5, 0.5) | **0.6593** | 0.8903 | 0.8605 |
| (0.95, 0.5, 0.5) | 0.6372 | 0.8998 | 0.8671 |
| (0.95, 0.95, 0.5) | 0.5174 | 0.9338 | 0.8892 |
| (0.5, 0.95, 0.5) | 0.6224 | 0.9111 | 0.8779 |
| (0.5, 0.5, 0.95) | 0.5911 | 0.9243 | 0.8891 |
| (0.95, 0.5, 0.95) | 0.5727 | 0.9281 | 0.8911 |
| (0.95, 0.95, 0.95) | 0.4696 | **0.9546** | **0.9139** |
| (0.5, 0.95, 0.95) | 0.5561 | 0.9432 | 0.9096 |
| (0.75, 0.75, 0.75) | 0.6114 | 0.9243 | 0.8924 |

**Figure 5.** Potential application of NeuroBridge for opportunistic cognitive impairment screening. (a) Schematic illustration of the proposed screening workflow. Brain MRI examinations acquired for routine clinical indications are processed using NeuroBridge classifiers for *AD vs EMCI*, *AD vs CN*, *LMCI vs CN*, and *EMCI vs CN*. Classifier probability thresholds are applied to stratify examinations into higher- and lower-risk groups and generate two screening outcomes: detection of *LMCI + AD vs EMCI + CN* and detection of all cognitively abnormal individuals (*EMCI, LMCI, and AD vs CN*). (b) Three-dimensional threshold analysis showing sensitivity, specificity, and positive predictive value (PPV) across classifier-threshold combinations. The top row presents results for all cognitively abnormal individuals versus CN, and the bottom row presents results for *LMCI + AD vs EMCI + CN*. (c) Representative operating points for selected threshold combinations, summarized using sensitivity, specificity, and PPV. Threshold triplets are ordered according to the classifiers specified in each table heading, and performance values are reported as proportions. Human illustrations are schematic and are not intended to represent individual study participants or the demographic composition of the study cohorts.

The threshold-response surfaces in Figure 5b demonstrate systematic threshold-dependent changes in screening performance across threshold combinations. Across most threshold settings, PPV remained above 80% for all abnormal screening and above 86% for LMCI + AD screening, while sensitivity and specificity varied according to the selected probability thresholds. In addition to screening performance, the computational efficiency of the framework was evaluated. Screening more than 1,000 MRI examinations required less than 7 minutes for each proposed screening scenario using a single RTX 4090 GPU, indicating that probability-based screening can be performed at scale with minimal computational overhead.

## 4 Discussion

The present study introduced NeuroBridge, a clinically guided multi-task MRI knowledge transfer framework for neurodegenerative disease diagnosis. By combining large-scale self-supervised MRI pretraining, task-specific anatomical supervision, and gated fusion fine-tuning, NeuroBridge consistently improved classification performance across multiple neurodegenerative diagnostic tasks and demonstrated strong robustness under cross-cohort evaluation. Across both ADNI and OASIS datasets, the framework outperformed directly trained, conventional 2D and 3D deep learning baselines in distinguishing AD, MCI-related populations, and cognitively normal controls. Furthermore, NeuroBridge produced informative probability analysis that enabled probability-threshold prediction and demonstrated the feasibility of opportunistic MRI-based screening strategies.

### 4.1 Multi-task representation learning and diagnostic performance

A major finding of this work is that preserving and integrating complementary MRI representations provides measurable advantages over conventional single-task fine-tuning. Neurodegenerative disease diagnosis rarely depends on a single imaging characteristic. Instead, clinicians integrate multiple structural observations, including anatomical localization, regional atrophy patterns, and overall structural integrity, when forming diagnostic impressions. NeuroBridge was designed to mimic this process computationally. Hippocampal segmentation provides information regarding anatomical boundaries and spatial localization, hippocampal atrophy classification captures disease-relevant degeneration patterns, and masked reconstruction preserves global anatomical organization. While each task individually captures only a subset of the information available within MRI, jointly leveraging these representations produces a richer description of brain structure. The multi-task pretraining results further demonstrate the feasibility of the proposed training framework. Prior to downstream fine-tuning, NeuroBridge successfully optimized all three auxiliary objectives, demonstrating accurate hippocampal localization, highly discriminative atrophy classification, and realistic anatomical reconstruction. Together, these results indicate that the model learned transferable neuroanatomical features spanning multiple levels of structural information and provided a robust initialization for subsequent dementia classification. The consistent downstream improvements observed across ADNI and OASIS further suggest that these representations contain complementary disease-relevant information that is not fully captured by conventional single-task learning approaches.

NeuroBridge improved performance across all evaluated classification tasks, including conventional *AD vs CN* diagnosis, achieving accuracies of 88.17% and 82.78% in ADNI and OASIS, respectively. Notably, the magnitude of improvement increased in diagnostically challenging settings involving mild cognitive impairment, early disease stages, and heterogeneous dementia populations. The largest gains were observed for *AD vs LMCI* (+8.74 percentage points over the strongest baseline), *CDR 0 vs CDR* 0.5 (+8.20 percentage points), and *AD + AD with Secondary Diagnosis vs CN* (+6.67 percentage points). These tasks represent transitions between neighboring disease states or mixed-pathology populations, where structural manifestations are often less consistent across individuals. The progressively larger gains observed as diagnostic complexity increases suggest that integrating anatomical localization, disease-related degeneration patterns, and global structural organization provides additional discriminative information beyond single-task learning. Clinically, this finding is particularly important because early-stage and heterogeneous cognitive impairment frequently present the greatest diagnostic challenges, while timely identification may provide greater opportunities for treatment planning, clinical monitoring, and trial enrollment.

## 4.2 Cross-cohort generalization, prediction probability, and screening implications

Another important finding is the strong cross-cohort generalization demonstrated by NeuroBridge. Robust transfer across independent cohorts remains a major challenge in medical AI because models frequently learn dataset-specific characteristics associated with imaging protocols, scanner hardware, population demographics, or diagnostic procedures (Zech et al., 2018). Despite these sources of variation, NeuroBridge consistently achieved the best performance when transferred between ADNI and OASIS without retraining. The improvements were particularly evident in recall and AUROC, suggesting enhanced preservation of disease sensitivity under distribution shift. Notably, the ADNI-trained NeuroBridge model achieved 84.71% accuracy on OASIS, exceeding the corresponding within-cohort OASIS performance of 82.78%. Although this comparison should be interpreted cautiously because the experimental settings are not identical, one possible explanation is that the larger sample size and more standardized MRI acquisition protocols of ADNI facilitate learning of representations that transfer more effectively across cohorts. Collectively, these findings indicate that the multi-task representations learned by NeuroBridge capture neuroanatomical characteristics that are less dependent on cohort-specific acquisition properties and more closely related to underlying disease biology.

The prediction probability analysis provides additional evidence regarding both representation quality and model reliability. Across both ADNI and OASIS cohorts, classification accuracy increased consistently as probability thresholds became more stringent, indicating that model-output probability was strongly associated with prediction correctness. Importantly, these improvements were not restricted to a small number of isolated cases. For example, ADNI *AD vs CN* accuracy increased from 91% to approximately 96% while retaining more than 85% of subjects, and OASIS *CDR 0 vs CDR 0.5* accuracy approached 96% while retaining nearly three quarters of the cohort. The largest probability-threshold-dependent gains were observed in MCI-related and heterogeneous dementia tasks, including *AD vs LMCI*, *CN vs EMCI*, *AD with Secondary Diagnosis vs CN*, and *Other Dementia vs CN* classifications. Together with the broader and more overlapping probability distributions observed in these tasks, this pattern suggests that prediction uncertainty appears to be associated with more ambiguous imaging patterns rather than arbitrary model behavior. Conversely, highly confident predictions correspond to more clearly separated disease phenotypes and progressively greater class separation within the probability distributions. Such behavior is particularly important in clinical environments, where AI systems are expected not only to generate predictions but also to communicate uncertainty. Probability-based deployment strategies may therefore allow highly reliable cases to be prioritized for automated assessment while directing lower-probability examinations toward additional expert review and clinical correlation (Kompa et al., 2021).

The opportunistic screening analysis extends the probability-based behavior of NeuroBridge to a potential clinical application. Opportunistic screening has gained increasing attention because it leverages imaging studies acquired for unrelated clinical indications to identify previously unrecognized diseases. In neurodegenerative disease, routine brain MRI examinations frequently contain information relevant to cognitive impairment, yet subtle structural abnormalities may not be the primary focus of image interpretation. To explore whether these imaging findings could support opportunistic screening, we evaluated the probability-based framework across a wide range of threshold combinations. The framework demonstrated smooth and predictable trade-offs between sensitivity, specificity, and PPV. At lower probability thresholds, the framework favored broader case detection, identifying up to 85.21% of cognitively abnormal individuals, whereas at higher thresholds, specificity and PPV approached 98%, indicating that positive screening outputs were highly enriched for true abnormalities. This flexibility suggests that the same underlying model can support different screening objectives, ranging from broad population-level surveillance to highly selective identification of individuals warranting further evaluation. Importantly, because the framework targets routinely acquired MRI examinations without requiring additional image acquisition, NeuroBridge could potentially be incorporated into prospective neuroimaging workflows to screen for atrophy patterns suggestive of cognitive impairment. Such implementation is also computationally feasible; in our experiments, screening more than 1,000 MRI examinations required less than 7 minutes for each screening scenario. Predictions can be generated in less than one second per scan, enabling results to be available at the time of image interpretation. Furthermore, class probabilities were obtained directly from the classifier softmax outputs without requiring additional uncertainty-modeling components, simplifying implementation and deployment. Although retrospective and exploratory, these findings illustrate a practical pathway for incorporating neurodegenerative risk assessment into routine clinical imaging workflows (Wenderott et al., 2024).

### 4.3 Limitations and future directions

This study also had several limitations to be considered. First, although NeuroBridge demonstrated favorable performance across ADNI and OASIS, additional validation on larger multi-center cohorts remains necessary. Both datasets represent highly curated research populations and may not fully reflect the heterogeneity encountered in routine clinical practice. Second, the current multi-task framework focuses primarily on hippocampal anatomy and hippocampal degeneration. While the hippocampus is a central biomarker of Alzheimer's disease and cognitive impairment, neurodegenerative disorders involve a broader range of structural abnormalities, including ventricular enlargement, cortical thinning, white matter injury, and distributed regional atrophy (Scheltens et al., 2002). Incorporating these complementary biomarkers may further strengthen diagnostic performance and biological interpretability. Third, the opportunistic screening analysis was retrospective and should not be interpreted as direct evidence of clinical effectiveness. Prospective studies will be required to further determine whether probability-based screening improves diagnostic workflows or patient outcomes. Finally, the present study focused primarily on cross-sectional structural MRI and did not incorporate longitudinal information. Disease progression trajectories and temporal imaging changes may provide additional predictive information beyond single time-point examinations.

Future work will focus on expanding the range of clinically informed auxiliary tasks, integrating multimodal imaging and non-imaging biomarkers, and incorporating longitudinal disease progression modeling. The framework is naturally extensible and could accommodate additional objectives related to white matter hyperintensities, cortical degeneration, ventricular expansion, microhemorrhages, and other neuroimaging markers associated with dementia. Furthermore, integration of cognitive assessments, genetic risk factors, fluid biomarkers, and longitudinal imaging may enable construction of unified multimodal representations capable of supporting both diagnosis and prognosis. Such developments may ultimately facilitate the creation of clinically informed neuroimaging foundation models that combine broad anatomical knowledge with disease-specific expertise.

Taken together, this work suggests that clinically guided multi-task representation learning provides an effective strategy for translating heterogeneous MRI-derived knowledge into robust diagnostic representations. By bridging self-supervised foundation-style pretraining with anatomically meaningful supervision and probability-based decision making, NeuroBridge offers a scalable framework for neurodegenerative disease assessment that remains effective across diverse datasets and diagnostic settings. This approach may provide a useful foundation for future AI systems designed not only for dementia classification but also for reliable clinical deployment and population-level neurodegenerative disease screening.

**Conflict of Interest**

*The authors declare that the research was conducted in the absence of any commercial or financial relationships that could be construed as a potential conflict of interest.*

**Author Contributions**

ML: Conceptualization, methodology, software, formal analysis, investigation, data curation, visualization, validation, and writing—original draft. GS: Methodology, software, data curation, validation, and writing—review and editing. CWF: Conceptualization, methodology, clinical interpretation, validation, investigation, data curation, supervision, project administration, funding acquisition, writing—original draft, and writing—review and editing. XZ: Conceptualization, methodology, software, formal analysis, investigation, data curation, visualization, validation, supervision, project administration, funding acquisition, writing—original draft, and writing—review and editing. All authors contributed to the article and approved the submitted version.

**Funding**

The author(s) declare that financial support was received for the research, authorship, and/or publication of this article. This work was supported by the Rajen Kilachand Fund for Integrated Life Science and Engineering and the Hariri Institute for Computing's Focused Research Programs (FRPs).

**Acknowledgments**

This work was supported by the Rajen Kilachand Fund for Integrated Life Science and Engineering and the Rafik B. Hariri Institute for Computing's Focused Research Programs (FRPs). We would like to thank the Boston University Photonics Center for technical support.

OpenAI ChatGPT (GPT-5.6 Sol; OpenAI, San Francisco, CA, United States) was used on a limited basis to assist with grammatical editing and to provide the general suggestions regarding the layout of the human-figure illustration in Figure 1. No AI-generated visual content was directly incorporated into any figure. All artwork, including the human illustrations, graphical elements, technical diagrams, labels, and figure layouts in the final submitted figures, was manually created by the authors. Generative AI was not used to generate, modify, analyze, or interpret scientific data, technical content, quantitative results, or conclusions. All scientific and technical information presented in the manuscript and figures was independently prepared and verified by the authors for accuracy. The original editable figure files (vector image documents) have been retained by the authors and are available upon request.

**NACC**
The NACC database is funded by NIA/NIH Grant U24 AG072122. NACC data are contributed by the NIA-funded ADRCs: P30 AG062429 (PI James Brewer, MD, PhD), P30 AG066468 (PI Oscar Lopez, MD), P30 AG062421 (PI Bradley Hyman, MD, PhD), P30 AG066509 (PI Thomas Grabowski, MD), P30 AG066514 (PI Mary Sano, PhD), P30 AG066530 (PI Helena Chui, MD), P30 AG066507 (PI Marilyn Albert, PhD), P30 AG066444 (PI David Holtzman, MD), P30 AG066518 (PI Lisa Silbert, MD, MCR), P30 AG066512 (PI Thomas Wisniewski, MD), P30 AG066462 (PI Scott Small, MD), P30 AG072979 (PI David Wolk, MD), P30 AG072972 (PI Charles DeCarli, MD), P30 AG072976 (PI Andrew Saykin, PsyD), P30 AG072975 (PI Julie A. Schneider, MD, MS), P30 AG072978 (PI Ann McKee, MD), P30 AG072977 (PI Robert Vassar, PhD), P30 AG066519 (PI Frank LaFerla, PhD), P30 AG062677 (PI Ronald Petersen, MD, PhD), P30 AG079280 (PI Jessica Langbaum, PhD), P30 AG062422 (PI Gil Rabinovici, MD), P30 AG066511 (PI Allan Levey, MD, PhD), P30 AG072946 (PI Linda Van Eldik, PhD), P30 AG062715 (PI Sanjay Asthana, MD, FRCP), P30 AG072973 (PI Russell Swerdlow, MD), P30 AG066506 (PI Glenn Smith, PhD, ABPP), P30 AG066508 (PI Stephen Strittmatter, MD, PhD), P30 AG066515 (PI Victor Henderson, MD, MS), P30 AG072947 (PI Suzanne Craft, PhD), P30 AG072931 (PI Henry Paulson, MD, PhD), P30 AG066546 (PI Sudha Seshadri, MD), P30 AG086401 (PI Erik Roberson, MD, PhD), P30 AG086404 (PI Gary Rosenberg, MD), P20 AG068082 (PI Angela Jefferson, PhD), P30 AG072958 (PI Heather Whitson, MD), P30 AG072959 (PI James Leverenz, MD).

**ADNI**
Data collection and sharing for this project was funded by the Alzheimer's Disease Neuroimaging Initiative (ADNI) (National Institutes of Health Grant U01 AG024904) and DOD ADNI (Department of Defense award number W81XWH-12-2-0012). ADNI is funded by the National Institute on Aging, the National Institute of Biomedical Imaging and Bioengineering, and through generous contributions from the following: AbbVie, Alzheimer’s Association; Alzheimer’s Drug Discovery Foundation; Araclon Biotech; BioClinica, Inc.; Biogen; Bristol-Myers Squibb Company; CereSpir, Inc.; Cogstate; Eisai Inc.; Elan Pharmaceuticals, Inc.; Eli Lilly and Company; EuroImmun; F. Hoffmann-La Roche Ltd and its affiliated company Genentech, Inc.; Fujirebio; GE Healthcare; IXICO Ltd.; Janssen Alzheimer Immunotherapy Research & Development, LLC.; Johnson & Johnson Pharmaceutical Research & Development LLC.; Lumosity; Lundbeck; Merck & Co., Inc.; Meso Scale Diagnostics, LLC.; NeuroRx Research; Neurotrack Technologies; Novartis Pharmaceuticals Corporation; Pfizer Inc.; Piramal Imaging; Servier; Takeda Pharmaceutical Company; and Transition Therapeutics. The Canadian Institutes of Health Research is providing funds to support ADNI clinical sites in Canada. Private sector contributions are facilitated by the Foundation for the National Institutes of Health ([www.fnih.org](http://www.fnih.org)). The grantee organization is the Northern California Institute for Research and Education, and the study is coordinated by the Alzheimer’s Therapeutic Research Institute at the University of Southern California. ADNI data are disseminated by the Laboratory for Neuro Imaging at the University of Southern California.

**OASIS**
Data in this study related to the OASIS repository were provided by OASIS 1-4:

OASIS-1: Cross-Sectional: Principal Investigators: D. Marcus, R, Buckner, J, Csernansky J. Morris; P50 AG05681, P01 AG03991, P01 AG026276, R01 AG021910, P20 MH071616, U24 RR021382

OASIS-2: Longitudinal: Principal Investigators: D. Marcus, R, Buckner, J. Csernansky, J. Morris; P50 AG05681, P01 AG03991, P01 AG026276, R01 AG021910, P20 MH071616, U24 RR021382

OASIS-3: Longitudinal Multimodal Neuroimaging: Principal Investigators: T. Benzinger, D. Marcus, J. Morris; NIH P30 AG066444, P50 AG00561, P30 NS09857781, P01 AG026276, P01 AG003991, R01 AG043434, UL1 TR000448, R01 EB009352. AV-45 doses were provided by Avid Radiopharmaceuticals, a wholly owned subsidiary of Eli Lilly.

OASIS-3_AV1451: Principal Investigators: T. Benzinger, J. Morris; NIH P30 AG066444, AW00006993. AV-1451 doses were provided by Avid Radiopharmaceuticals, a wholly owned subsidiary of Eli Lilly.

OASIS-4: Clinical Cohort: Principal Investigators: T. Benzinger, L. Koenig, P. LaMontagne

## Supplementary Material

The Supplementary Material for this article is available as a separate file.

## Data Availability Statement

The **ADNI** dataset used in the current study is available on the Alzheimer's Disease Neuroimaging Initiative office website at: https://adni.loni.usc.edu/. To access the ADNI data repository, one must register with a valid account and complete the online application process.

The **NACC** dataset used in the current study is available on the National Alzheimer's Coordinating Center office website at: https://naccdata.org/requesting-data/data-request-process. To access the NACC data repository, registration with a valid account is required, followed by the online application process.

The **OASIS** dataset used in the current study can be applied in the Open Access Series of Imaging Studies office website at: https://sites.wustl.edu/oasisbrains/. To access the OASIS data repository, registration with a valid account is required, followed by the online application process.

The **RadImageNet** dataset used in the current study can be accessed by creating an account on the RAD IMAGE NET official website at: https://www.radimagenet.com/. To access the RadImageNet data repository, registration with a valid account is required, followed by the online application process.

Derived results, trained models, and analysis scripts are available from the corresponding author upon request.